\documentclass[letterpaper, 10 pt, conference]{ieeeconf}  
\IEEEoverridecommandlockouts                              
\usepackage[T1]{fontenc}
\usepackage[utf8]{inputenc}
\usepackage{subfiles}
\usepackage[cmex10]{amsmath}              
\usepackage{graphicx}
\usepackage{subcaption}
\usepackage{epstopdf}
\usepackage{amssymb}                            
\usepackage{mdwmath}                           
\usepackage{threeparttable}                     
\usepackage[usenames, dvipsnames]{color}
\usepackage{empheq}
\usepackage{cancel}
\usepackage{soul}
\usepackage{colortbl}
\usepackage{array, indentfirst}
\usepackage{multirow}
\usepackage{siunitx}
\usepackage{bm}
\usepackage{algorithm}
\usepackage{algpseudocode}
\usepackage{xcolor}
\algdef{SE}[VARIABLES]{Variables}{EndVariables}
   {\algorithmicvariables}
   {\algorithmicend\ \algorithmicvariables}
\algnewcommand{\algorithmicvariables}{\textbf{global variables}}
\algdef{SE}[GIVEN]{Given}{EndGiven}
{\algorithmicgiven}
   {\algorithmicend\ \algorithmicgiven}
\algnewcommand{\algorithmicgiven}{\textbf{given}}
\algdef{SE}[GIVEN]{Hypers}{EndHypers}
{\algorithmichypers}
   {\algorithmicend\ \algorithmichypers}
\algnewcommand{\algorithmichypers}{\textbf{hyperparameters}}
\newcolumntype{P}[1]{>{\centering\arraybackslash}p{#1}}
\newcolumntype{M}[1]{>{\centering\arraybackslash}m{#1}}

\title{
\LARGE \bf
Motion-Specific Battery Health Assessment for Quadrotors Using High-Fidelity Battery Models
}

\author{
Joonhee Kim$^{*1}$, Sanghyun Park$^{*1}$, Donghyeong Kim$^{2}$, Eunseon Choi$^{1}$, and Soohee Han$^{**1}$% <-this % stops a space
\thanks{*The first two authors contributed equally to this work.}% <-this % stops a space
\thanks{**Corresponding author.}% <-this % stops a space
\thanks{$^{1}$Authors are with the Department of Convergence IT Engineering,
        Pohang University of Science and Technology, 
        Choengam-ro 77, Nam-gu, Pohang-si, Gyeongsangbuk-do, Republic of Korea.
        {\tt\small \{junhemeike, pash0302, eunseon103, sooheehan\}@postech.ac.kr}}%
\thanks{$^{2}$Author is with the Department of Electrical Engineering,
        Pohang University of Science and Technology, 
        Choengam-ro 77, Nam-gu, Pohang-si, Gyeongsangbuk-do, Republic of Korea.
        {\tt\small dongh5290@postech.ac.kr}}%
}

\begin{document}
\maketitle
\thispagestyle{empty}
\pagestyle{empty}

%%%%%%%%%%%%%%%%%%%%%%%%%%%%%%%%%%%%%%%%%%%%%%%%%%%%%%%%%%%%%%%%%%%%%%%%%%%%%%%%
\begin{abstract}

Quadrotor endurance is ultimately limited by battery behavior, yet most energy-aware planning treats the battery as a simple energy reservoir and overlooks how flight motions induce dynamic current loads that accelerate battery degradation. 
This work presents an end-to-end framework for motion-aware battery health assessment in quadrotors. 
We first design a wide-range current sensing module to capture motion-specific current profiles during real flights, preserving transient features. 
In parallel, a high-fidelity battery model is calibrated using reference performance tests and a metaheuristic based on a degradation-coupled electrochemical model.
By simulating measured flight loads in the calibrated model, we systematically resolve how different flight motions translate into degradation modes—loss of lithium inventory and loss of active material—as well as internal side reactions.
The results demonstrate that even when two flight profiles consume the same average energy, their transient load structures can drive different degradation pathways, emphasizing the need for motion-aware battery management that balances efficiency with battery degradation.

\end{abstract}
%%%%%%%%%%%%%%%%%%%%%%%%%%%%%%%%%%%%%%%%%%%%%%%%%%%%%%%%%%%%%%%%%%%%%%%%%%%%%%%%

\section{Introduction}
The industrial adoption of quadrotors—for applications such as aerial inspection, logistics, and environmental monitoring—has significantly accelerated in recent years, driven by advances in autonomy and maneuverability~\cite{floreano2015science}.
Despite their versatility, quadrotor endurance remains fundamentally constrained by battery performance.
That is, the capability and reliability of onboard batteries directly determine mission feasibility, payload capacity, and safety margins.
To tackle such constraints, quadrotor research has advanced battery/energy-aware strategies aimed at enabling versatile mission profiles, enhancing autonomy, and ensuring reliable performance in demanding applications.

Early research integrated energy constraints into quadrotor planning and control.
For example, by explicitly minimizing energy rather than path length, or by determining the speed that minimizes energy expenditure per unit distance~\cite{morbidi2016minimum, di2015energy, datsko2024energy}.
Subsequent studies extended these efforts to higher-level objectives such as multi-agent exploration, autonomous battery swapping, and so on~\cite{cesare2015multi, barrett2018autonomous, choudhry2021cvar, seewald2024energy}.
Although such approaches enhanced endurance and operational safety, they typically treated the battery as a simple energy reservoir rather than as an electrochemical system.
This abstraction overlooks a practical reality: Quadrotor flight motions generate time-varying current profiles, and even at identical average energy, transient features—peak amplitude, ripple, and duty cycle—can affect batteries in different ways~\cite{uddin2016effects, goldammer2022impact, paw2023battery, al2025energy}.
Consequently, a clear need has emerged to quantify how motion-specific current profiles drive battery consumption and degradation.

Research on quadrotor motion-battery interaction spans from motion-specific power estimation to direct measurements of motion-specific degradation. 
In previous studies~\cite{abeywickrama2018comprehensive, jacewicz2022quadrotor, gong2023modeling}, closed-form multi-rotor models and empirical profiling forecast the energy required for forward flight, climbs, descents, and turns, but they largely stop at consumption.
More recent studies push further by imposing mission segments or mining flight logs to quantify how maneuver-induced loads shorten endurance and accelerate degradation~\cite{paw2023battery, al2025energy}.
These contributions provide valuable evidence—often at the module/system scale or via black-box predictors—yet they rarely resolve which transient features of the current waveform trigger specific degradation mechanisms.
As a result, a physical bridge from motion-specific current profiles to solid-electrolyte interphase (SEI) growth, lithium plating, and loss of active material remains missing.
This gap motivates electrochemical degradation analysis for dynamic current profiles induced by various motions in real quadrotor flights. 

Here, we establish an end-to-end framework for motion-aware battery health assessment in quadrotors.
A custom wide-range sensing module is developed to capture motion-specific current profiles during real flights, preserving transient features such as peak amplitudes, ripple, and duty cycles.
These measured profiles are then used to calibrate a high-fidelity battery model, constructed from a degradation-coupled electrochemical model and parameterized via reference performance tests with a metaheuristic optimization algorithm.
By replaying the measured currents in the calibrated model, we resolve how distinct maneuvers translate into degradation modes—loss of active material (LAM) and loss of lithium inventory (LLI)—as well as internal side reactions such as SEI layer growth and particle cracking.
In doing so, we expose cases where identical average energy yields markedly different degradation trajectories due to the waveform structure of the current load, bridging the gap between energy consumption metrics and electrochemical and mechanistic degradation insights.

The main contributions of this paper are as follows:
\begin{itemize}
    \item An end-to-end framework that integrates real flight sensing, battery model calibration, and virtual experiments for motion-aware battery health assessment.
    \item A high-fidelity, degradation-coupled battery electrochemical model enabling virtual experiments, where realistic flight loads are reproduced and mechanistic degradation pathways are resolved.
    \item A systematic analysis connecting quadrotor flight motion with internal degradation pathways to guide motion-aware battery management.
\end{itemize}

The remainder of this paper is organized as follows.
Section II introduces the degradation-coupled electrochemical model and parameter estimation framework.
Section III describes the proposed framework, including the sensing module, battery model calibration, and health assessment indicators.
Section IV presents experimental results from quadrotor flights, highlighting how motion-specific profiles impact degradation relative to constant-current baselines.
Finally, Section V concludes the paper.

% Measurement: We design and deploy a 0–100 A custom current-sensing module to acquire maneuver-annotated flight current with sufficient temporal detail to preserve transient features.

% Digital twin calibration: We perform 0.1 C RPT with a custom cycler and identify a P2D model whose parameters reproduce the reference behavior, providing a physics-based twin for degradation analysis under arbitrary flight profiles.

% Mechanism-aware attribution and metrics: We replay measured profiles in the twin to quantify mechanism-specific responses (SEI growth rate, plating onset/accumulation proxies, and active-material loss indicators) and distill them into motion-aware health metrics (e.g., damage-per-energy/distance and maneuver-to-damage sensitivity). These metrics inform trajectory design, operating envelopes, and BMS limits, paving the way for damage-aware (not merely energy-aware) decision making in UAV operations.

\section{Battery Electrochemical Model}

To accurately capture the high C-rate discharge currents--often exceeding 2 C-rate and reaching high instantaneous peaks--observed in quadrotors, it is insufficient to rely on the equivalent circuit models or simplified electrochemical models like the single particle model (SPM).
While the SPM offers computational advantages, it frequently neglects electrolyte-phase dynamics and concentration gradients, which become non-negligible and lead to significant voltage prediction errors under the aggressive load profiles typical of UAVs.
Instead, high-fidelity electrochemical models are required to maintain predictive accuracy across the full operational range.
Among them, the pseudo-two-dimensional (P2D) model offers a balanced compromise between physical fidelity and computational tractability~\cite{plett2015battery}. 
By employing porous electrode theory, it achieves computational times on the order of seconds, making it feasible for iterative simulation. 
At the same time, the P2D model resolves microscopic electrochemical dynamics with high fidelity under dynamic and high current conditions. 
Beyond short-term dynamics, high-rate operation also accelerates degradation: Steep concentration gradients and elevated overpotential inside the electrodes exacerbate SEI thickening, lithium plating, and loss of active material. 
To capture such effects, the P2D framework has been extensively extended with advanced sub-models for degradation and fault processes~\cite{ren2018investigation,kong2020pseudo}.

Building on this foundation, we adopt a state-of-the-art degradation-coupled P2D model, which integrates SEI layer growth, lithium plating, and particle cracking~\cite{o2022lithium}.
To account for the thermal effects of high-rate discharge, a lumped thermal model is also incorporated into the framework.
This combination allows the model to capture not only the effects of individual side reactions but also their coupled impacts, which intensify under dynamic current profiles. 
In practice, quadrotor batteries frequently experience high-rate discharges during flights, leading to elevated temperatures and mechanical stress. 
These physical changes accelerate both electrochemical and mechanical side reactions.
By explicitly representing such mechanisms, the adopted model provides an effective platform to evaluate how motion-specific current profiles translate into detailed degradation pathways.
%, thereby supporting motion-aware battery health assessment.

\begin{table}[!t]
\renewcommand{\tabcolsep}{0.2cm} 
\renewcommand{\arraystretch}{1.25} \small
    \centering
    \caption{Battery model parameters to be estimated}
    \begin{tabular}{p{3.5cm}p{4cm}}
        \hline\hline
        \multicolumn{2}{l}{Electrochemical parameters} \\ \hline
        Symbol & Description \\ \hline
        $\theta_p^{\text{max}}, \theta_p^{\text{min}}, \theta_n^{\text{max}}, \theta_n^{\text{min}}$&Stoichiometry limit  \\
        $R_p,R_n$&Solid particle radius \\
        $\sigma_p, \sigma_n$&Solid-phase conductivity \\
        $\epsilon_p,\epsilon_{\text{sep}},\epsilon_n$&Electrolyte porosity \\
        $\epsilon_{f,p},\epsilon_{f,n}$&Filler fraction \\
        $a_p,a_n$&Specific surface area\\
        $D_e,D_p,D_n$&Diffusivity \\
        $c_{e}^{\text{init}}$&Initial lithium concentration\\
        $i_{0,p}^\text{ref},i_{0,n}^\text{ref}$&Reference current density\\ \hline
        \multicolumn{2}{l}{Degradation parameters} \\ \hline
        Symbol & Description \\ \hline
        $R_{\text{SEI}}$&SEI layer resistance\\ 
        $\nu_{p}, \nu_{n}$ & Poisson's ratio \\
        $E_{p}, E_{n}$ & Young's modulus \\
        $V_{p}, V_{n}, V_{\text{Li}}$ & Partial molar volume \\
        \hline\hline
        \multicolumn{2}{l}{$\star$ \textit{p} and \textit{n} refer to the cathode and anode, respectively.} \\
        \multicolumn{2}{l}{$\star$ \text{sep}, \textit{e}, and init refer to the separator, electrolyte, and} \\
        \multicolumn{2}{l}{\quad the initial value, respectively.} \\
    \end{tabular}
    \label{table:param}
\end{table}

\begin{figure*}[!t]
    \centering
    \includegraphics[width=0.95\textwidth]{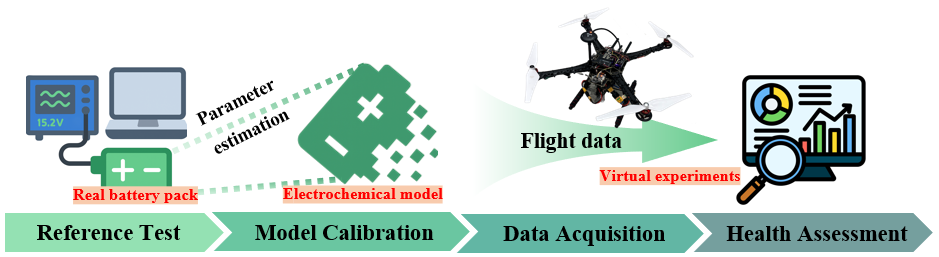}
    \caption{
    Pipeline of motion-specific battery health assessment: reference tests calibrate an electrochemical model, flight data are acquired with a custom module, and virtual experiments evaluate degradation.
    }
    \label{fig:pipeline}
\end{figure*}

The adopted model represents a real battery cell using only a few dozen electrochemical and degradation parameters, as summarized in Table~\ref{table:param}.
These parameters are identified by solving an optimization problem that aligns the model’s voltage response with experimental measurements under prescribed current profiles.
To this end, we employ metaheuristic optimization algorithms, which iteratively adjust parameter values to minimize the root-mean-square error (RMSE) between simulated and measured voltage responses:
\begin{equation}
\vec{\theta}_{\text{best}} = \underset{\vec{\theta}}{\operatorname{argmin}} \sqrt{\frac{1}{N}\sum_{t=1}^N \big \{V_{\text{real}}(\vec{c},t) - V_{\text{P2D}}(\vec{\theta},\vec{c},t)\big\}^2},
\label{eqn:criterion}
\end{equation}
where $\vec{\theta}$ denotes the parameter vector in Table~\ref{table:param}, $N$ is the number of sampled time points, $\vec{c}$ represents the operating conditions (including input current and temperature), and $V_{\text{real}}$ and $V_{\text{P2D}}$ are the measured and simulated voltage responses, respectively.
Metaheuristic approaches are particularly well suited to this task, as they efficiently explore high-dimensional, non-convex parameter spaces and converge to physically consistent solutions without requiring gradient information.

% Quadrotor의 높은 C-rate의 방전 전류를 모사하기 위해선 로보틱스 분야에서 널리 사용되는 등가회로모델이 아닌 고정밀의 전기화학 모델을 사용해야한다.
% Pseudo-two-dimensional (P2D) 모델은 microscopic dynamics 기반의 물리 모델로 고전류에 대해 높은 정확도를 가지면서도 porous electrode theory를 사용한 연속체 모델로서 몇 분 정도의 tolerable한 계산 속도를 가진다. 
% 이런 잘알려진 P2D 모델은  다양한 열화 현상 및 fault에 대한 advanced sub-model들에 대한 연구가 잘되어있으며, 이번 연구에서는 최신의 열화 결합 P2D 모델을 사용한다.

% P2D 모델은 오직 십 수개의 물리 및 전기화학적 파라미터 만으로 실제 배터리를 표현할 수 있는데, 일반적으로 샘플 기반 최적화 기법인 메타 휴리스틱 알고리즘을 사용하여 목표 배터리 응답을 모사하도록 모델 파라미터를 추정한다.
% 배터리 모델 파라미터를 최적화는 구체적으로 목표 배터리의 실제 전압, 전류 데이터를 수집하여 배터리 모델로 하여금 같은 전류를 입력하였을 때 전압 응답을 실제와 같이 맞추는 과정이다.
% 이를 위해, 그림과 같이 battery in the loop system을 구성하여 0.1 C-rate 및 2 C-rate 로 reference performance test를 진행하여 실제 데이터를 수집하였다.
% 이와 같이 C-rate를 설정한 이유는 실제 비행 전류가 고전류인 것을 고려하여 저전류와 고전류에서 다르게 응답하는 배터리의 특성을 가상 배터리에 모두 담아내고자 함이다.
\section{Motion-Specific Battery Health Assessment}

Figure~\ref{fig:pipeline} illustrates the overall pipeline of the proposed motion-specific battery health assessment framework. 
The pipeline begins with reference performance tests (RPTs) on the target battery pack, providing voltage–current data for parameter estimation.
Based on the reference data, we estimate battery model parameters to calibrate a high-fidelity battery electrochemical model that accurately reproduces the dynamic responses of the target pack.
Flight data are then collected from quadrotor experiments using a custom sensing module capable of capturing wide-range discharge currents in real time.
Finally, the measured motion-specific current profiles are replayed in the calibrated model to assess how different motions activate distinct degradation pathways.
The following subsections describe each component of this pipeline: Battery model calibration, flight data acquisition, and motion-specific battery health assessment.

\subsection{Battery model calibration}

\begin{figure}[!t]
    \centering
    \includegraphics[width=0.40\textwidth]{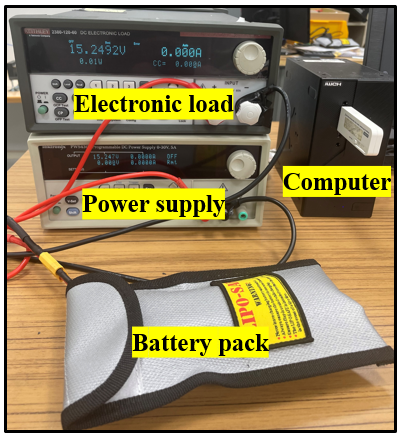}
    \caption{
    Experimental setup of the battery-in-the-loop system, consisting of a programmable power supply, an electronic load, and a battery pack under test.
    }
    \label{fig:BILS}
\end{figure}

To obtain the experimental data required for parameter estimation, we first implement a battery-in-the-loop system (BILS), as illustrated in Figure~\ref{fig:BILS}. 
The BILS integrates a programmable power supply, an electronic load, and a host controller, allowing the real battery pack to be operated under controlled charge–discharge profiles while its voltage and current responses are simultaneously recorded. 
Using this setup, we conduct RPTs at 0.1 and 2 C-rate to capture both low-rate equilibrium behavior and high-rate dynamic behavior, thereby providing calibration data across the operating regimes most relevant to quadrotor applications.

The acquired datasets are then used to estimate the electrochemical and degradation parameters of the P2D model. For this purpose, we employ strategically switching metaheuristics (SSM), a state-of-the-art metaheuristic algorithm that adaptively optimizes the parameters in response to the problem landscape to achieve efficient and accurate calibration~\cite{kim2023strategically}.
The effectiveness of SSM has been validated across diverse battery chemistries, demonstrating robustness and reliability in capturing internal dynamics. 
This broad validation confirms that the same approach is well suited to calibrating the quadrotor battery pack considered in this study.
Furthermore, SSM has been successfully applied in prior work to enable virtual experiments, where calibrated battery models reproduce realistic battery responses with high fidelity. 
This capability makes SSM particularly suitable for the present work, where dynamic and motion-specific current profiles from real flights can be reliably reproduced and analyzed within the calibrated battery model.

\begin{figure}[!t]
    \centering
    \includegraphics[width=0.48\textwidth]{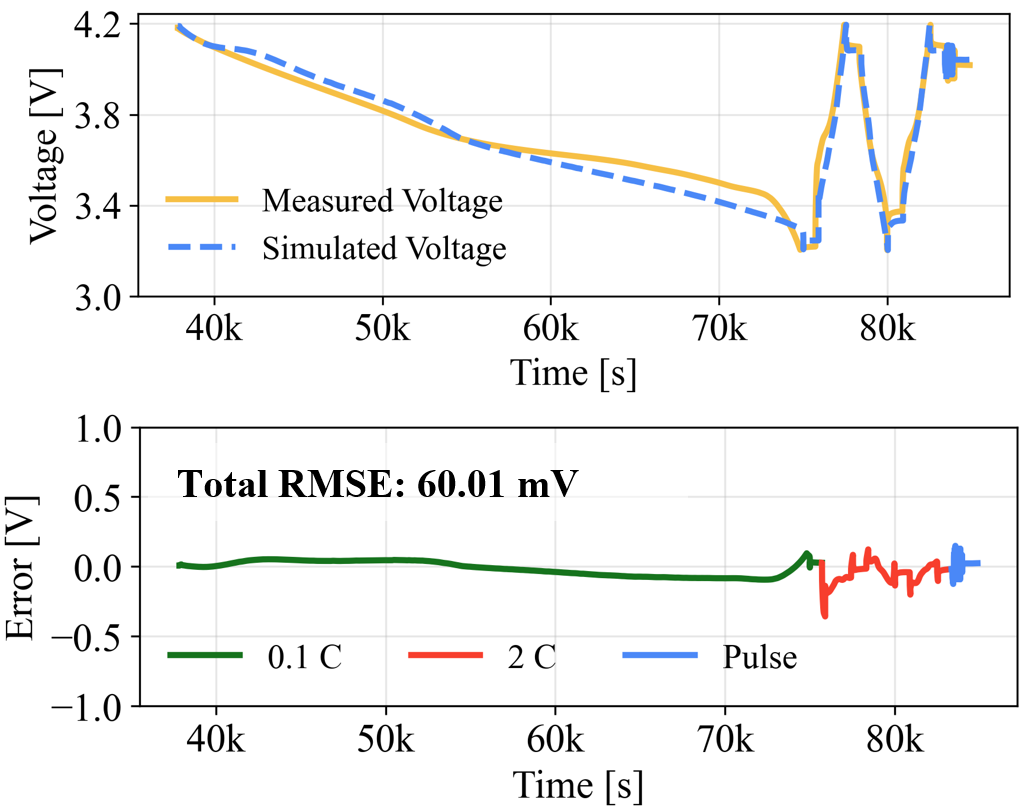}
    \caption{
    Calibration results of the battery electrochemical model. (Top) Comparison between measured and simulated voltage responses under 0.1 C, 2 C, and pulse reference performance tests. (Bottom) Corresponding error profile, with segments distinguished by load conditions.
    }
    \label{fig:esti}
\end{figure}
% which presents the measured and simulated voltage responses alongside their corresponding error profiles under the RPT protocols.

Figure~\ref{fig:esti} illustrates the calibration effectiveness, comparing measured and simulated responses under the RPT protocols.
The calibrated model closely reproduces the measured voltage trajectories across low-rate, high-rate, and dynamic pulse conditions, effectively capturing both steady-state and transient characteristics. 
In practice, quadrotor battery packs typically consist of multiple cells connected in series.
In this study, we assume that all cells are uniformly manufactured and that inter-cell resistance is negligible, allowing the measured pack voltage to be represented by multiplying the model voltage by the number of series-connected cells.
This simplification enables the use of a single-cell electrochemical model to represent the pack behavior.
Under this assumption, the residual error remains consistently small, resulting in a total RMSE of only 60.01 mV.
These results confirm that the calibrated battery model accurately captures both steady-state and transient behaviors, providing a reliable foundation for the subsequent motion-specific health assessments.

\subsection{Flight data acquisition}

\begin{figure}[!t]
    \centering
    \includegraphics[width=0.48\textwidth]{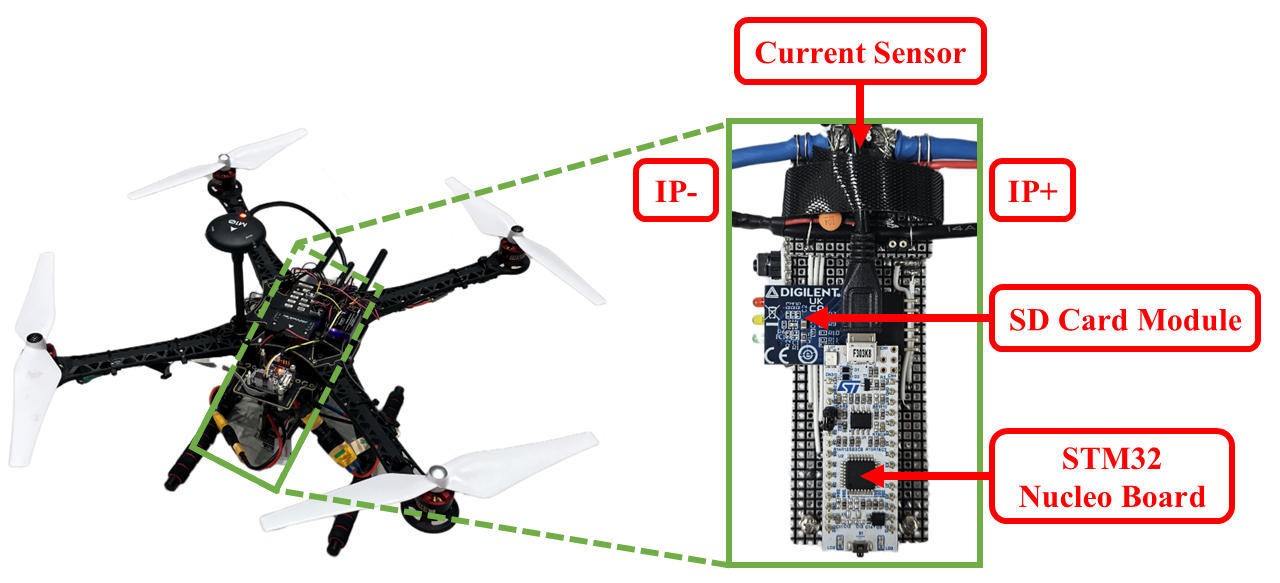}
    \caption{
    Custom current sensing module integrated into the quadrotor platform. The module consists of a wide-range current sensor, an STM32 Nucleo board for control and data processing, and an SD card module for onboard data logging.
    }
    \label{fig:SensorModule}
\end{figure}

\begin{table}[!t]
\renewcommand{\tabcolsep}{0.2cm} 
\renewcommand{\arraystretch}{1.25} \small
    \centering
    \caption{Detailed specifications of the quadrotor platform and the 4S1P battery pack employed in this study.}
    \begin{tabular}{p{2.2cm}p{5.5cm}}
        \hline\hline
        \multicolumn{2}{l}{Quadrotor specification} \\ \hline
        Parameter & Description \\ \hline
        Frame & S500 Quadrotor  \\
        Controller & Pixhawk 6C \\
        Motor & DJI NAZA 2212 Brushless Motor 920KV \\
        Propeller & DJI 1045 propeller \\
        Weight & 975 g \\ \hline
        \multicolumn{2}{l}{Battery pack specification} \\ \hline
        Parameter & Description \\ \hline
        Capacity & 2.2 Ah \\
        Voltage & 14.8 V \\
        Max C-rate & 35 \\
        Weight & 235 g \\
        Configuration & 4S1P \\
        \hline\hline
    \end{tabular}
    \label{table:spec}
\end{table}

Data acquisition was conducted using a custom quadrotor platform integrated with a specialized current sensing module, as illustrated in Figure~\ref{fig:SensorModule} and Table~\ref{table:spec}.
The flight control system utilizes a cascaded proportional-integral-derivative architecture consisting of an outer position loop and an inner attitude loop, with desired velocities commanded via a remote control transmitter during experiments.
To evaluate battery impact, the flight tests were designed around three distinct motions: Hovering, vertical maneuvering, and horizontal translation.
While quadrotor dynamics are traditionally decoupled into translational and rotational motions for underactuated control design~\cite{taeyoung2010Geomtery}, this study categorizes these motions based on force direction and power consumption.
From a battery diagnostics perspective, distinguishing between vertical, aligned with the thrust vector, and horizontal motion, orthogonal to gravity, provides more physically meaningful insights into degradation mechanisms~\cite{abeywickrama2018comprehensive, jacewicz2022quadrotor}.

To accurately capture the high-magnitude current profiles typical of quadrotor flight (ranging between 10 and 100 A), the sensing module employs a $\pm$ 100 A wide-range sensor.
This allows for precise measurement of both nominal and aggressive peak loads.
Consistent with prior quadrotor data acquisition studies~\cite{MILLER2024104262,Rodrigues2021}, the current is sampled at 10 Hz. Each timestamped measurement is logged to an onboard SD card, providing synchronized datasets for subsequent analysis.

\begin{figure}[!t]
    \centering
    \includegraphics[width=0.45\textwidth]{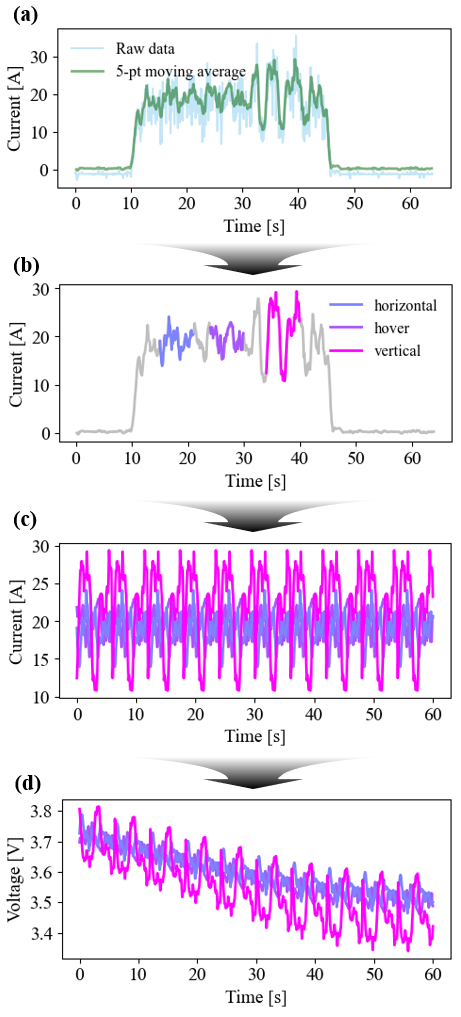}
    \caption{
    Procedure of extraction and processing of motion-specific current profiles from quadrotor flight data. (a) Raw current signal with a moving-average filter. (b) Segmentation of the filtered data. (c) Periodic reconstruction of the segmented profiles. (d) Simulated voltage responses under the reconstructed profiles.
    }
    \label{fig:current}
\end{figure}

\begin{figure*}[!t]
    \centering
    \includegraphics[width=0.95\textwidth]{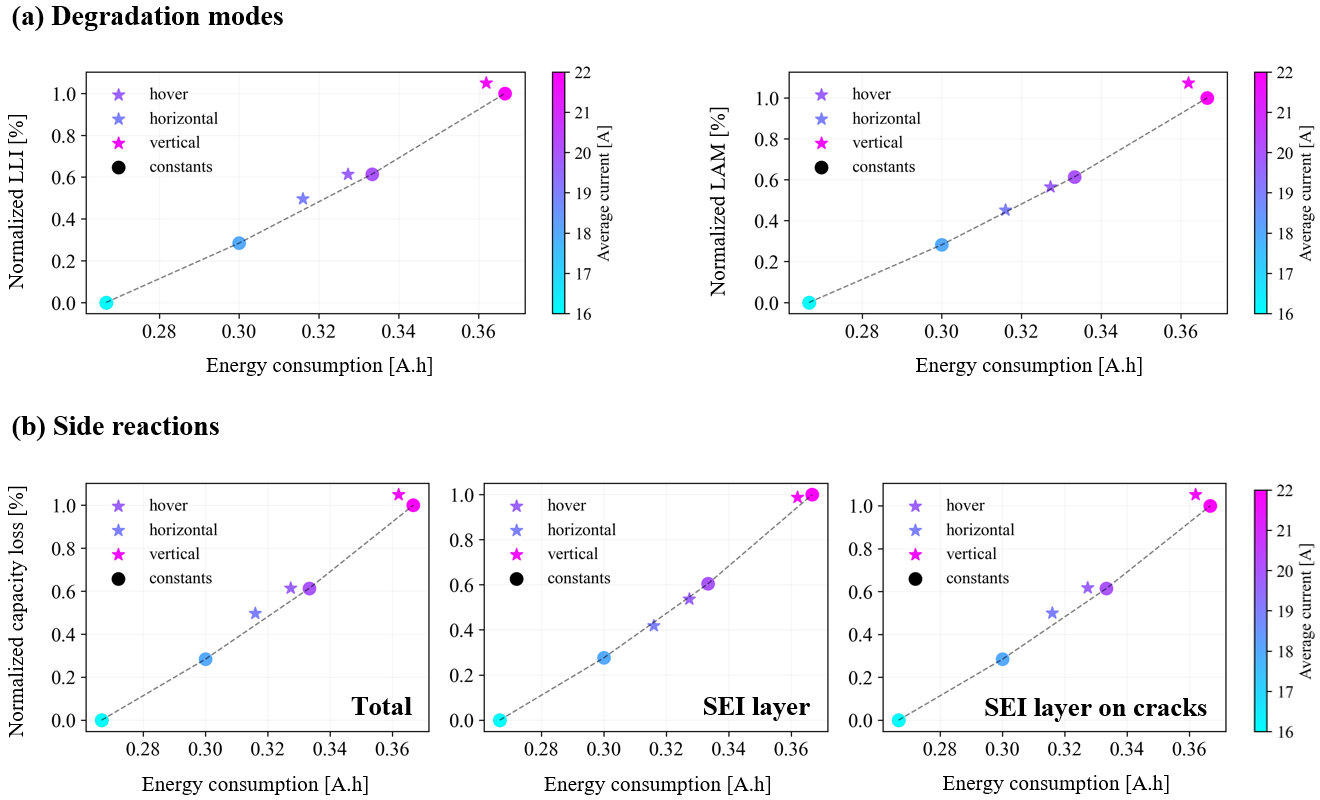}
    \caption{
    Motion-aware battery health assessment results under motion-specific (stars) and constant-current (circles) profiles with C-rates of 16, 18, 20, and 22. (a) Normalized degradation modes (b) Normalized capacity loss induced by side reactions.
    % In (b), the near-identical trends between 'Total' and 'SEI layer on cracks' indicate that this specific side reaction plays a dominant role in the overall capacity loss.
    }
    \label{fig:results}
\end{figure*}

\subsection{Battery health assessment on calibrated battery model}

With the calibrated battery model and real flight profiles, the framework proceeds to health assessment, quantifying how different flight motions drive distinct battery degradation.
The process begins by replaying segmented flight profiles in the calibrated battery model, which incorporates the degradation sub-models described in Section 2.
The model then provides both cell-level performance indicators and degradation mechanism-specific insights, offering a comprehensive perspective of quadrotor battery health.

The assessment considers two levels of degradation phenomena: Degradation modes and side reactions.
At the mode level, we evaluate loss of lithium inventory (LLI) and loss of active material (LAM)~\cite{han2019review}. 
LLI represents the gradual depletion of cyclable lithium, primarily consumed by side reactions such as SEI growth or lithium plating.
LAM corresponds to the irreversible deactivation of electrode regions, often caused by particle cracking or surface isolation.
Together, these two modes contribute to overall capacity fade. 
In practical battery packs, however, measurement of LAM and LLI requires time-intensive \textit{ex-situ} experiments, which may themselves risk inducing unintended battery damage~\cite{pastor2017comparison}.

At the reaction level, the degradation-coupled battery model enables quantification of capacity loss induced by SEI layer growth, lithium plating, and particle cracking.
These side reactions represent competing degradation pathways, and the extent to which each affects the future trajectory of battery health.
For example, SEI layer growth consumes cyclable lithium and reduces coulombic efficiency; while its rate diminishes as the layer thickens, dynamic quadrotor operation can expose fresh surfaces that sustain renewed growth.
Particle cracking, by contrast, progresses slowly at first but accelerates over time, promoting structural deterioration and loss of active material.
The advanced model used in this study further captures SEI formation on newly generated cracks, reflecting the coupled nature of these processes. 
By diagnosing the balance among such mechanisms, the framework not only quantifies total capacity loss but also identifies the electrochemical pathways that dictate long-term degradation risks under motion-specific current profiles.

\section{Experimental Results}
\subsection{Motion-specific current profile extraction}

Figure~\ref{fig:current} illustrates the procedure for extracting and processing motion-specific current profiles from quadrotor flight experiments. 
Figure~\ref{fig:current}(a) shows raw current data alongside a moving-average filter used to smooth measurement noise.
Based on this averaged profile, distinct segments corresponding to specific motions—such as hovering, vertical motion, and horizontal motions—are identified and color-coded as shown in Figure~\ref{fig:current}(b).
This segmentation isolates the current characteristics of each motion from the mixed flight trajectory, enabling systematic comparison across motion types.
The segmented profiles are then reconstructed into periodic waveforms, as shown in Figure~\ref{fig:current}(c), allowing each motion’s current signature to be repeatedly replayed within the calibrated battery model.
Finally, Figure~\ref{fig:current}(d) presents the corresponding simulated voltage responses, which capture how different current profiles influence cell-level dynamics.
Thus, the extracted and standardized profiles serve as controlled inputs to the calibrated battery model, forming the basis for the motion-specific health assessment presented in the following section.

\subsection{Motion-aware battery health assessment}

To highlight motion-specific degradation effects, the assessment includes constant current baselines at 16 A and 22 A, which cover average current values, i.e., the average energy consumptions, of the motion-specific profiles.
All degradation metrics are normalized with respect to these baselines, so that deviations directly indicate the additional impact induced by each flight motion.
As shown in Figure~\ref{fig:results}, the baselines (circles) exhibit a nearly linear increase in degradation with energy consumption, consistent with the expectation that higher C-rates proportionally accelerate battery degradation.

At the mode level in Figure~\ref{fig:results}(a), LLI increases more steeply than the baselines for all motions, showing that dynamic current profiles consistently accelerate lithium consumption. 
In contrast, LAM exhibits a pronounced nonlinear rise only under the vertical profile, while hover and horizontal motions remain close to the linear baseline trend.
At the reaction level in Figure~\ref{fig:results}(b), the LLI trend corresponds closely to the total contribution of side reactions, confirming that lithium loss is primarily driven by parasitic processes. 
Among the components, SEI layer growth on nominal surfaces stays near the baseline except for the vertical case, where greater heat generation amplifies the reaction. 
SEI growth on cracks, however, is elevated across all motions, reflecting the universal formation of micro-cracks under dynamic current profiles and their role in sustaining parasitic reactions.

Synthesizing the mode- and reaction-level results, the degradation process under quadrotor flight can be interpreted as a two-stage pathway. 
First, all motions induce micro crack formation, which sustains additional SEI growth on newly exposed surfaces and accelerates lithium loss. 
This outcome is consistent with the small fluctuations observed in the hover and horizontal profile, which repeatedly perturb the electrodes and promote crack initiation without causing significant material isolation. 
Second, only vertical motion exhibits the large-amplitude peaks and frequent transients evident in their current profile, generating sufficient thermal and mechanical stress to propagate cracks and disconnect electrode regions. 
This condition surpasses the threshold for active material isolation, producing a marked increase in LAM. 
Consequently, while hover and horizontal motions primarily intensify lithium inventory loss through crack-mediated side reactions, vertical motion drives a fundamentally different degradation regime in which electrochemical and mechanical pathways act in concert, resulting in disproportionately severe overall aging.

Taken together, these results demonstrate that the battery degradation is not determined solely by total energy consumption.
While constant-current baselines imply a simple and rate-dependent scaling, motion-specific current profiles introduce nonlinear effects through both electrochemical and mechanical pathways. 
In practice, this means that flight control strategies optimized only for energy efficiency may inadvertently accelerate aging: Aggressive motions, although energy-efficient in the short term, impose thermal and mechanical stresses that amplify particle cracking and parasitic reactions.
These findings highlight the need for motion-aware battery management, where motion planning accounts for both energy use and the underlying degradation mechanisms.
\section{Discussion}
\subsection{Discussion on alternative health diagnostic approaches}
The proposed physics-based framework provides a mechanistic depth that purely data-driven methods often lack. 
Recent studies have demonstrated the power of advanced machine learning and deep learning architectures in predicting battery state of health (SOH)~\cite{madani2025comparative, jiang2026battery}.
However, these approaches typically rely on extensive empirical datasets and function as "black-box" predictors that do not resolve specific internal degradation pathways.
To bridge this gap, emerging physics-informed neural networks have begun to co-estimate SOH and short-term degradation paths by embedding physical constraints into deep learning models~\cite{yang2025physics}. 
While these hybrid models improve interpretability, the P2D approach remains superior for quadrotor applications as it explicitly resolves the electrolyte-phase dynamics and concentration gradients that are critical under the aggressive, high-rate discharge profiles typical of flight.

\begin{table}[!t]
\renewcommand{\tabcolsep}{0.2cm} 
\renewcommand{\arraystretch}{1.25} \small
    \centering
    \caption{Specification of the secondary quadrotor platform}
    \begin{tabular}{p{2.2cm}p{5.5cm}}
        \hline\hline
        \multicolumn{2}{l}{Quadrotor specification} \\ \hline
        Parameter & Description \\ \hline
        Frame & Custom carbon frame (arm length 300mm)  \\
        Controller & Custom \\
        Motor & EMAX ECO2 2306 1700KV \\
        Propeller & Gemfan 7040-3 7inch \\
        Weight & 455 g \\ \hline
        \hline\hline
    \end{tabular}
    \label{table:nova}
\end{table}

\begin{figure}[!t]
    \centering
    \includegraphics[width=0.48\textwidth]{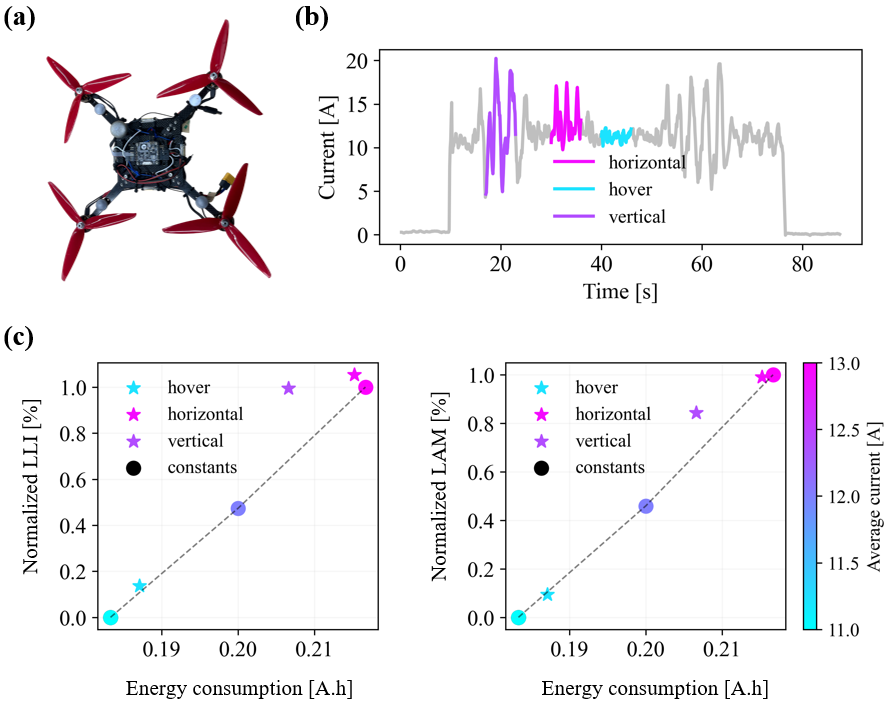}
    \caption{
    Health assessment results for a secondary 455 g quadrotor platform. (a) Hardware configuration. (b) Segmented motion-specific current profiles. (c) Normalized degradation modes. 
    }
    \label{fig:nova}
\end{figure}

\subsection{Toward generalization across diverse quadrotor platforms}
While a comprehensive statistical validation across a vast fleet of quadrotors is a subject for future research, the generalization of the proposed framework was explored using a secondary, lightweight quadrotor platform with distinct propulsion characteristics equipped with the same battery pack (Table~\ref{table:nova}).
This additional test serves to provide preliminary indications of whether the mechanistic link between flight dynamics and battery aging remains consistent across different hardware scales.
As illustrated in Figure~\ref{fig:nova}(a) and (b), motion-specific current profiles were successfully captured and processed from this secondary platform.

The health assessment results for this platform, presented in Figure~\ref{fig:nova}(c), further support the hypothesis that transient load structures are as significant as average energy consumption in driving battery degradation.
Interestingly, while the vertical motion did not consume the highest level of average energy in this specific configuration, it consistently induced the most significant non-linear degradation.
This observation corroborates the mechanistic conclusions established in the primary results specifically that the high-amplitude peaks inherent in vertical motions trigger non-linear degradation pathways regardless of the total energy throughput.
By further validating that these transient characteristics exacerbate mechanical and electrochemical aging across different hardware scales, these results reinforce the findings from the initial experiments and underscore the robustness of the proposed motion-aware health assessment framework.

\section{Conclusion}

This paper presented an end-to-end framework for motion-specific battery health assessment in quadrotors. 
A degradation-coupled electrochemical model was first calibrated through reference performance tests using a battery-in-the-loop system. 
In parallel, a custom wide-range sensing module was integrated into a quadrotor platform to capture motion-specific current profiles during real flights. 
By replaying these profiles in the calibrated model, we systematically linked flight motions to both degradation modes and underlying side reactions.

The experimental analysis revealed that constant-current baselines produce a nearly linear scaling of degradation with energy consumption, whereas motion-specific profiles introduce distinct nonlinearities. Vertical maneuvers, in particular, imposed large peaks and frequent transients in their current profiles, driving particle cracking beyond the isolation threshold and amplifying SEI growth on cracks. Hover and horizontal motions, by contrast, primarily intensified lithium inventory loss through crack-mediated side reactions without triggering significant active material loss. These results highlight that transient load features—not just average energy consumption—critically govern the trajectory of battery degradation.

Overall, this study demonstrates the importance of integrating flight dynamics into battery analysis and establishes a foundation for motion-aware battery management. 
Future work will extend this framework to diverse quadrotor mission scenarios and investigate integration with online management systems for real-time endurance and safety optimization.

\addtolength{\textheight}{-8cm}   % This command serves to balance the column lengths


% Please add this appendix section

\section*{ACKNOWLEDGMENT}
This work was supported in part by the Information Technology Research Center (ITRC) supervised by IITP under Grant IITP-2026-RS-2024-00438430.

The authors used ChatGPT and Gemini during the preparation of this manuscript. Specifically, Gemini was utilized to generate several image components featured in Figure~\ref{fig:pipeline}, while both ChatGPT and Gemini were employed for language editing and grammar enhancement throughout the entire manuscript. Following the use of these tools, the authors reviewed and edited the content as needed and take full responsibility for the integrity and accuracy of the publication.

% Please add this acknowledgment section

%%%%%%%%%%%%%%%%%%%%%%%%%%%%%%%%%%%%%%%%%%%%%%%%%%%%%%%%%%%%%%%%%%%%%%%%%%%%%%%%

\bibliography{ref}

\end{document}